\documentclass[10pt,twocolumn,letterpaper]{article}

\usepackage[]{cvpr}      

\usepackage[dvipsnames]{xcolor}

\definecolor{darkred}{rgb}{0.7, 0.0, 0.0}
\definecolor{darkgreen}{rgb}{0.0, 0.37, 0.14}
\definecolor{darkblue}{rgb}{0.10, 0.17, 0.8}

\definecolor{cvprblue}{rgb}{0.21,0.49,0.74}
\usepackage[pagebackref,breaklinks,colorlinks,citecolor=cvprblue]{hyperref}

\def\paperID{844} 
\def\confName{CVPR}
\def\confYear{2024}

\title{Selective-Stereo: Adaptive Frequency Information Selection for Stereo Matching}

\author{Xianqi Wang\footnotemark[1], ~~Gangwei Xu\footnotemark[1], ~~Hao Jia, ~~Xin Yang\footnotemark[2]\\
[2mm]
Huazhong University of Science and Technology
\\
{\tt\small \{xianqiw, gwxu, haojia, xinyang2014\}@hust.edu.cn}
}

\begin{document}
\maketitle
\begin{abstract}
Stereo matching methods based on iterative optimization, like RAFT-Stereo and IGEV-Stereo, have evolved into a cornerstone in the field of stereo matching. However, these methods struggle to simultaneously capture high-frequency information in edges and low-frequency information in smooth regions due to the fixed receptive field. As a result, they tend to lose details, blur edges, and produce false matches in textureless areas. In this paper, we propose Selective Recurrent Unit (SRU), a novel iterative update operator for stereo matching. The SRU module can adaptively fuse hidden disparity information at multiple frequencies for edge and smooth regions. To perform adaptive fusion, we introduce a new Contextual Spatial Attention (CSA) module to generate attention maps as fusion weights. The SRU empowers the network to aggregate hidden disparity information across multiple frequencies, mitigating the risk of vital hidden disparity information loss during iterative processes. To verify SRU's universality, we apply it to representative iterative stereo matching methods, collectively referred to as Selective-Stereo. Our Selective-Stereo ranks $1^{st}$ on KITTI 2012, KITTI 2015, ETH3D, and Middlebury leaderboards among all published methods. Code is available at \textcolor{magenta}{https://github.com/Windsrain/Selective-Stereo}.
\end{abstract}  

\renewcommand{\thefootnote}{\fnsymbol{footnote}}
\footnotetext[1]{Equal contribution.}
\footnotetext[2]{Corresponding author.}

\section{Introduction}
\label{sec:intro}

Stereo matching is a fundamental area of research in computer vision. It explores the calculation of displacement, referred to as disparity, between matching points in a pair of rectified images. This technique plays a significant role in various applications, including 3D reconstruction and autonomous driving.

With the advancement of deep learning, learning-based stereo matching~\cite{kendall2017end, chang2018pyramid, guo2019group, xu2022accurate, lipson2021raft, xu2023iterative} have progressively displaced traditional methods and significantly enhancing the accuracy of disparity estimation. Initially, aggregation-based methods~\cite{zhang2019ga, xu2022attention, cheng2022region} led the development of stereo matching algorithms. These methods begin by defining a maximum range of disparity, constructing a 4D cost volume using feature maps, and subsequently employing 3D CNN to filter the volume and derive the final disparity map. Such methods focus on filtering the initially coarse cost volume, thus effectively aggregating geometry information. However, cost aggregation requires a large number of convolutions, resulting in high computational costs, making it difficult to be applied to high-resolution images.

\begin{figure}[t]
    \centering
    \begin{subfigure}{1.0\linewidth}
        \begin{subfigure}{0.49\linewidth}
            \centering
            \includegraphics[width=1.0\linewidth]{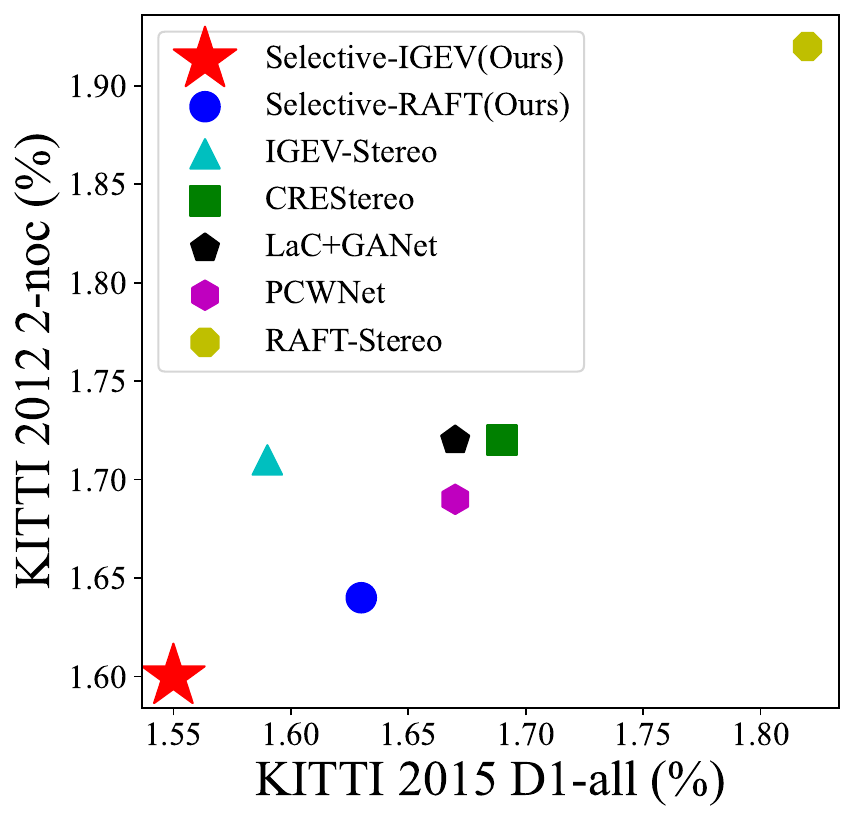}
        \end{subfigure}
        \begin{subfigure}{0.47\linewidth}
            \centering
            \includegraphics[width=1.0\linewidth]{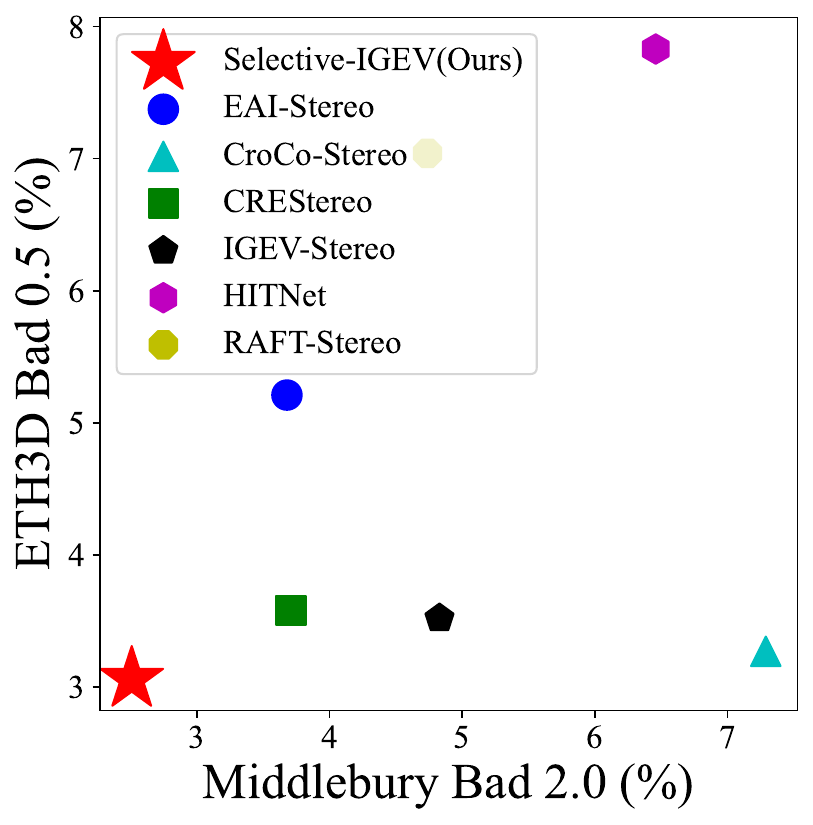}
        \end{subfigure}
        \vspace{5px}
    \end{subfigure}
    \begin{subfigure}{1.0\linewidth}
        \begin{subfigure}{0.32\linewidth}
            \centering
            \includegraphics[width=1.0\linewidth]{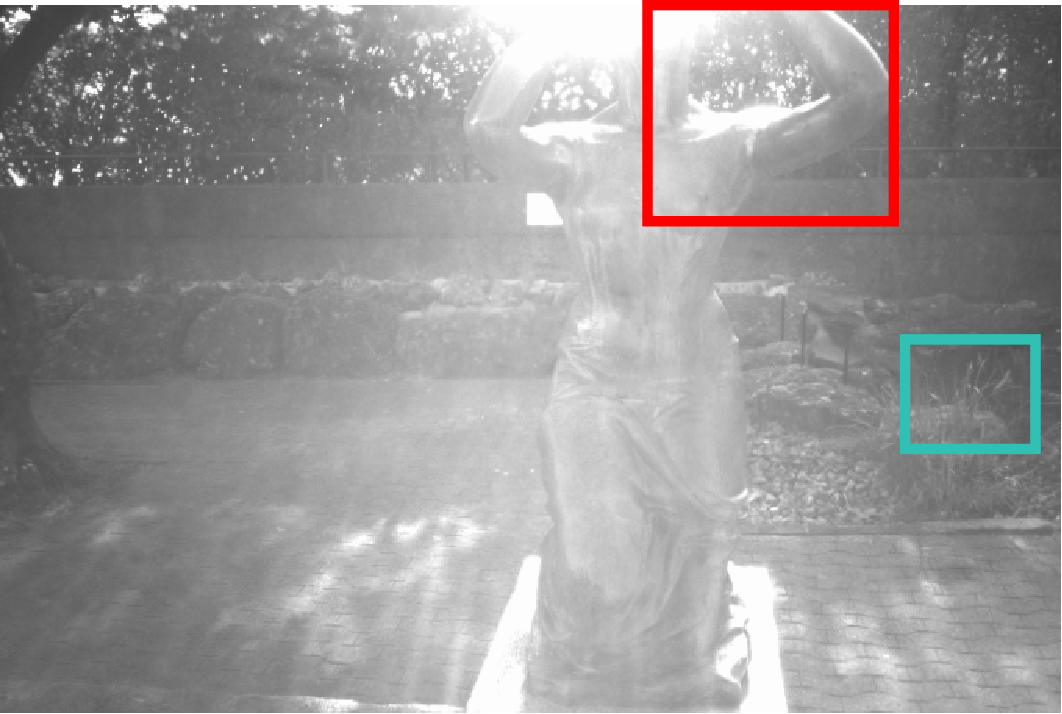}
            \caption*{Left Image}
        \end{subfigure}
        \begin{subfigure}{0.32\linewidth}
            \centering
            \includegraphics[width=1.0\linewidth]{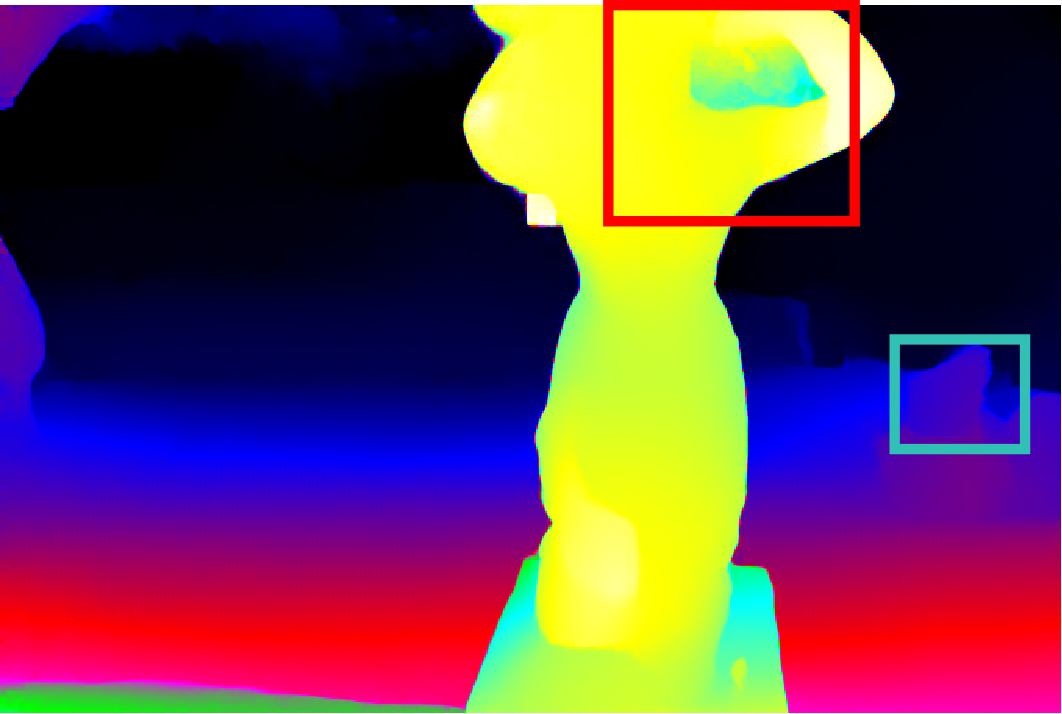}
            \caption*{RAFT-Stereo~\cite{lipson2021raft}}
        \end{subfigure}
        \begin{subfigure}{0.32\linewidth}
            \centering
            \includegraphics[width=1.0\linewidth]{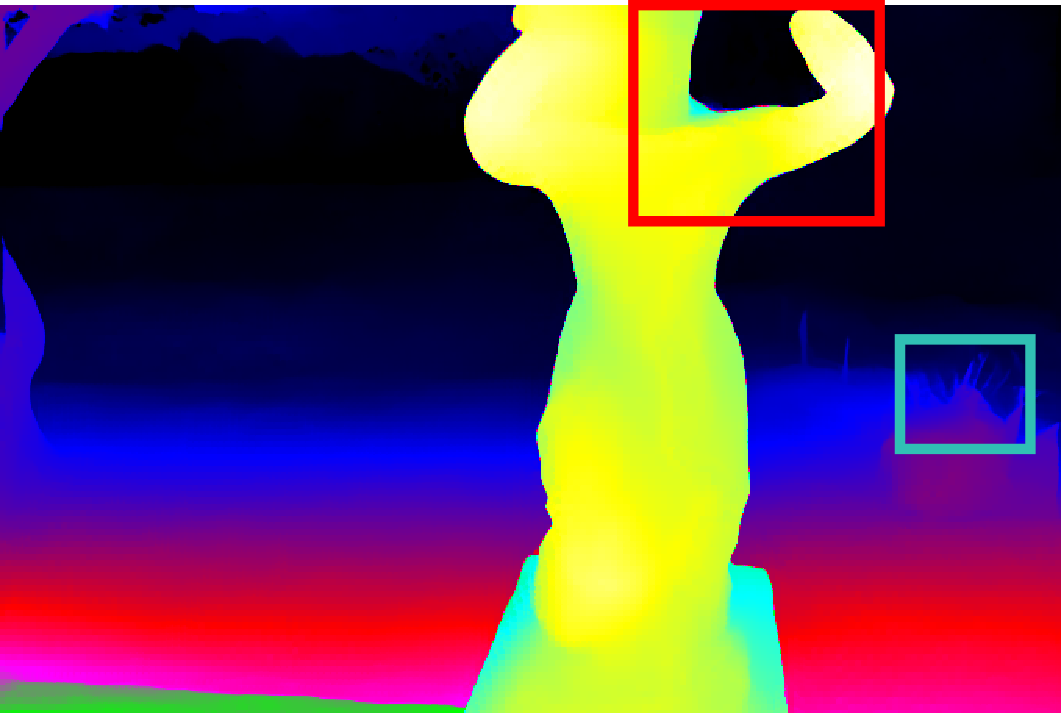}
            \caption*{Selective-RAFT}
        \end{subfigure}
        \vspace{5px}
    \end{subfigure}
    \begin{subfigure}{1.0\linewidth}
        \begin{subfigure}{0.32\linewidth}
            \centering
            \includegraphics[width=1.0\linewidth]{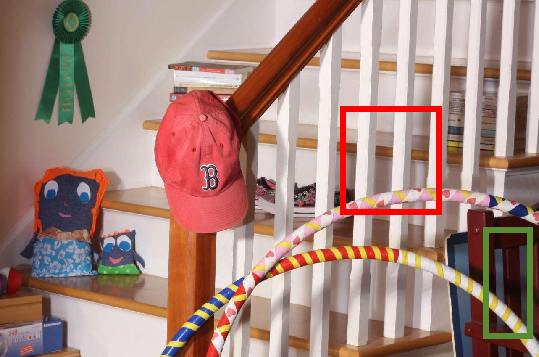}
            \caption*{Left Image}
        \end{subfigure}
        \begin{subfigure}{0.32\linewidth}
            \centering
            \includegraphics[width=1.0\linewidth]{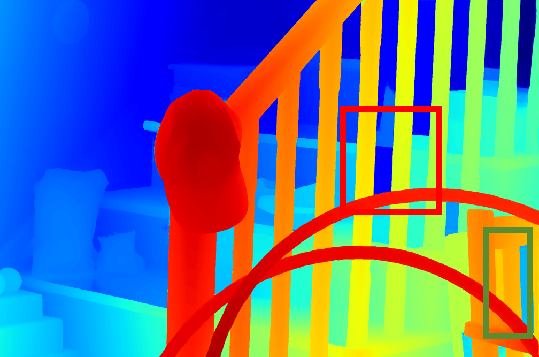}
            \caption*{IGEV-Stereo~\cite{xu2023iterative}}
        \end{subfigure}
        \begin{subfigure}{0.32\linewidth}
            \centering
            \includegraphics[width=1.0\linewidth]{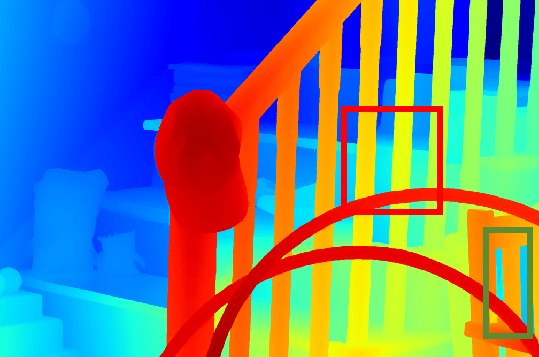}
            \caption*{Selective-IGEV}
        \end{subfigure}
        \vspace{-10pt}
    \end{subfigure}

    \caption{\textbf{Row 1:} Comparisons with state-of-the-art stereo methods on KITTI 2012~\cite{geiger2012we} and KITTI 2015~\cite{menze2015object}, ETH3D~\cite{schops2017multi} 
 and Middlebury~\cite{scharstein2014high} leaderboards. \textbf{Row 2:} Visual comparison with RAFT-Stereo on ETH3D. \textbf{Row 3:} Visual comparison with IGEV-Stereo on Middlebury. Our method distinguishes subtle details and sharp edges and performs well in weak texture regions.}
    \label{fig:ranking}
\vspace{-10pt}
\end{figure}

Recently, a novel class of methods based on iterative optimization~\cite{li2022practical, lipson2021raft, zhao2023high, feng2023mc} has been gaining prominence and achieving state-of-the-art performance on several leaderboards. These methods begin by constructing an all-pairs cost volume, indexing a local cost volume from the original cost volume, and subsequently employing recurrent units~\cite{cho2014learning} to calculate disparity residuals and update the disparity prediction. One major advantage of these methods is their ability to capture all candidate matching points without predefining the range of disparities. Additionally, these methods don't need to aggregate the cost volume using a large number of redundant convolutions. Instead, a continuous update of the disparity prediction is achieved through lightweight recurrent units during iterations. Therefore, these methods are capable of processing high-resolution images.

However, iterative methods encounter several challenges. Firstly, the all-pairs cost volume includes considerable noisy information~\cite{xu2023iterative}, potentially causing the loss of crucial information when iterating the hidden information. Besides, as the network iterates, the hidden information increasingly incorporates global low-frequency information while losing local high-frequency information like edges and thin objects~\cite{zhao2023high}. Secondly, the existing recurrent units possess a fixed receptive field, leading the network to solely concentrate on information at the current frequency and ignore other frequencies, such as detailed, edge, and textureless information.

In this paper, we propose Selective Recurrent Unit (SRU) to address the limitations of traditional recurrent units. As Chen \etal~\cite{chen2019drop} mentions features contain information at different frequencies, high-frequency information describes rapidly changing fine details, while low-frequency information describes smoothly changing structures. Unlike traditional recurrent units that treat information at different frequencies equally, our SRU incorporates multiple branches of GRU, each with a distinct kernel size representing different receptive fields. The hidden information obtained from each GRU branch is fused and then fed into the next iteration. This fusion enables the capture of information from different receptive fields at different frequencies, while also performing secondary filtering to reduce noise information from local cost volume. To further enhance the fusion process, we propose a Contextual Spatial Attention (CSA) module to utilize the context information. Instead of simply summarizing information from different branches, CSA introduces attention maps extracted from the context information. After doing so, information captured by small kernels has large weights in regions like edge, while information captured by large kernels has large weights in regions like low-texture. These attention maps determine the weight of fusion, allowing the network to adaptively select suitable information based on different image regions. Besides, we prove the effectiveness and universality of our module by transferring it to different iterative networks. All networks are collectively referred to as Selective-Stereo. By doing so, we consistently improve the performance of these networks without introducing a significant increase in parameters and time.

We demonstrate the effectiveness of our method on several stereo benchmarks. On Scene Flow~\cite{mayer2016large}, our Selective-RAFT reaches the state-of-the-art EPE of 0.47, and our Selective-IGEV even achieves a new state-of-the-art EPE of 0.44. And as shown in Fig. \ref{fig:ranking}, our Selective-RAFT surpasses RAFT-Stereo by a large margin and achieves competitive performance compared with the state-of-the-art methods on KITTI~\cite{geiger2012we, menze2015object} leaderboards. Our Selective-IGEV ranks $1^{st}$ on KITTI, ETH3D~\cite{schops2017multi}, and Middlebury~\cite{scharstein2014high} leaderboards among all published methods.

Our main contributions can be summarized as follows: 
\begin{itemize}
\item We propose a novel iterative update operator SRU for iterative stereo matching methods.
\item We introduce a new Contextual Spatial Attention module that generates attention maps for adaptively fusing hidden disparity information at multiple frequencies.
\item We verify the universality of our SRU on several iterative stereo matching methods.
\item Our method outperforms existing published methods on public leaderboards such as KITTI, ETH3D, and Middlebury.
\end{itemize}

\section{Related Work}
\label{sec:relate}

\textbf{Aggregation-based methods in stereo matching.} Several aggregation-based methods~\cite{kendall2017end, chang2018pyramid, guo2019group, xu2022accurate, xu2023cgi, cheng2022region, cheng2023coatrsnet, zhang2019ga, xu2020aanet, shen2022pcw} have shown significant progress in the domain of stereo matching in recent years. DispNet~\cite{mayer2016large} establishes the groundwork for subsequent network architecture. GC-Net~\cite{kendall2017end} proposes a 4D concatenate cost volume, which is subsequently regularized using 3D CNNs. Additionally, it also introduces the soft argmin function for disparity regression, resulting in a significant influence on subsequent methods. PSMNet~\cite{chang2018pyramid} proposes a stacked hourglass 3D CNN, which improves the cost aggregation stage to enhance the network's ability to capture context information. GwcNet~\cite{guo2019group} proposes Group-wise Correlation Volume, which combines the advantages of correlation and concatenation volume. GA-Net~\cite{zhang2019ga} designs a semi-global guided aggregation layer and a local guided aggregation layer, inspired by the traditional semi-global matching algorithm, to further assist in aggregating global and geometry information in the network. Building upon Group-wise Correlation Volume, ACVNet~\cite{xu2022attention} proposes Attention Concat Volume, which uses attention weights to suppress redundant information and maintain sufficient information for matching.

\begin{figure*}[t]
    \centering
    \includegraphics[width=1.0\textwidth]{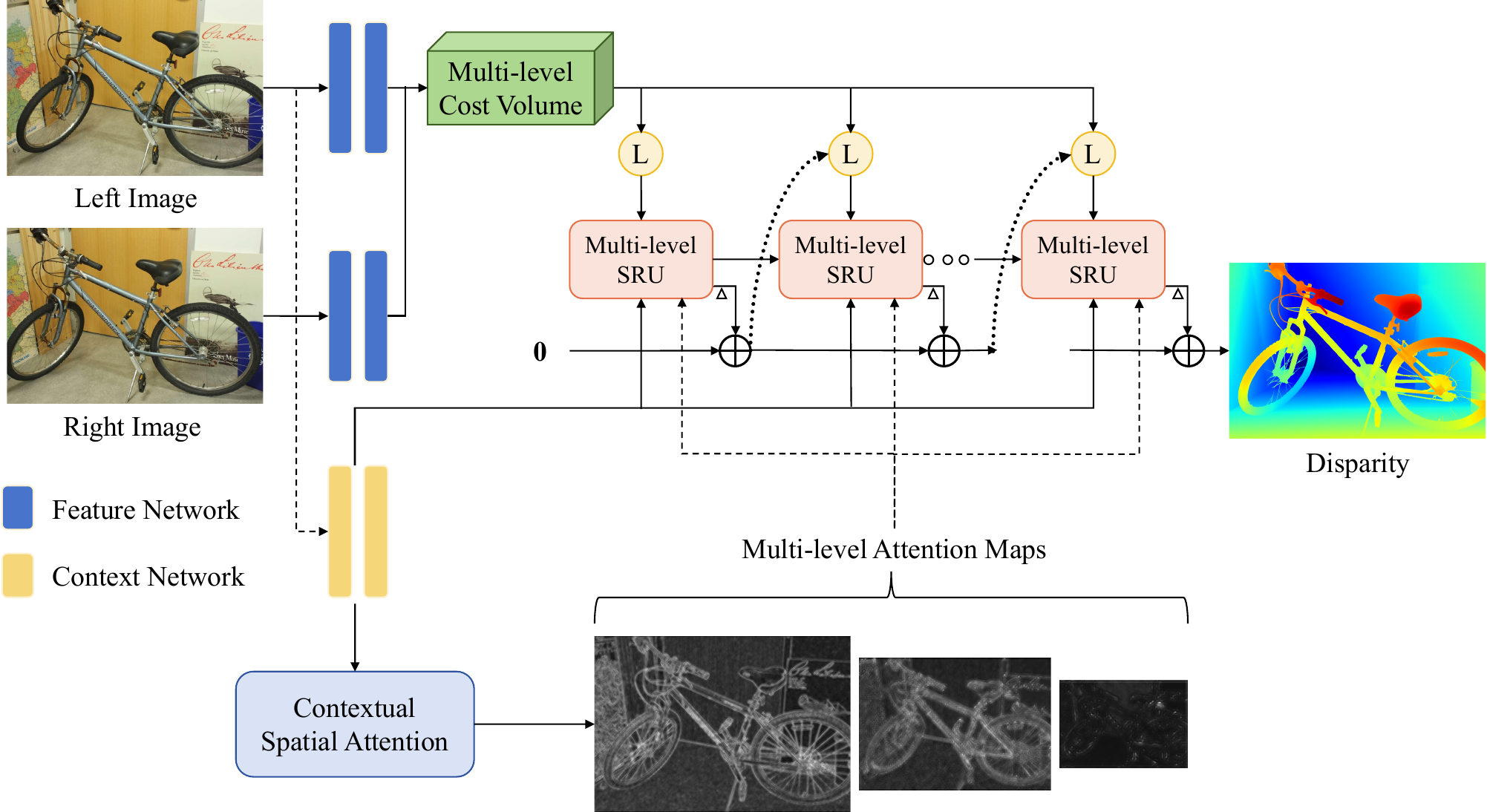}
    \caption{Overview of our proposed Selective-Stereo (Selective-RAFT version). The Contextual Spatial Attention (CSA) module extracts attention maps from context information as a guide for Selective Recurrent Units (SRUs). Then the network iteratively updates the disparity using local cost volumes retrieved from the correlation pyramid and attention maps given by CSA through SRUs.}
    \label{fig:overall}
    \vspace{-10pt}
\end{figure*}

\textbf{Iterative-based methods in stereo matching.} In recent years, many iterative methods~\cite{teed2020raft,lipson2021raft,xu2023iterative,xu2023memory}, spearheaded by RAFT~\cite{lipson2021raft}, have gradually become the mainstream of research. RAFT-Stereo~\cite{lipson2021raft} builds upon the optical flow method RAFT~\cite{teed2020raft} by introducing an all-pairs cost volume pyramid that maintains high resolution. It extracts local correlation features from this pyramid, performs iterative disparity updates using GRU-based update operators, and incorporates a multi-level GRU to expand the receptive field. On this basis, IGEV-Stereo~\cite{xu2023iterative} asserts that the initial cost volume is excessively coarse. To alleviate the need for iterations and reduce time overhead, it proposes to use a lightweight cost aggregation network before iterations. CREStereo~\cite{li2022practical} designs a hierarchical network in a coarse-to-fine manner, as well as a stacked cascaded architecture for inference in place of the original single-resolution iterative structure. DLNR~\cite{zhao2023high} proposes the use of LSTM as a replacement for GRU, providing the advantage of decoupling the update of hidden states from disparity prediction.

\textbf{Frequency information application in vision}. There are several works that focus on using frequency information in computer vision. Chen \etal~\cite{chen2019drop} propose the octave convolution to factorize the mixed feature maps by their frequencies. Xu \etal~\cite{xu2020learning} propose a method of learning in the frequency domain and suggest that CNN models are more sensitive to low-frequency channels than high-frequency. DSGAN~\cite{fritsche2019frequency} introduces the frequency separation into super-resolution. LITv2~\cite{pan2022fast} proposes to disentangle the high/low-frequency patterns in an attention layer.
\section{Method}
\label{sec:methods}

In this section, we present the overall architecture of Selective-Stereo. 
Because our method can be plugged into different networks, we take Selective-RAFT (Fig. \ref{fig:overall}) as an example and focus on illustrating its key components.

\subsection{Feature Extraction}
To ensure fair comparisons, Selective-RAFT maintains consistency with RAFT-Stereo~\cite{lipson2021raft} by employing its feature extraction network. Feature extraction comprises two main components: feature network and context network.

\textbf{Feature Network.} Given the left and the right images $\mathbf{I}_{l(r)} \in \mathbb{R}^{3 \times H \times W}$, we first downsample them to $1/2$ resolution using a $7 \times 7$ convolutional layer. Then, a series of residual blocks is employed to extract features and we apply another downsampling layer to get features at $1/4$ resolution in the middle. Lastly, a $1 \times 1$ convolutional layer is applied to get the final left and right features $\mathbf{f}, \mathbf{g} \in \mathbb{R}^{C \times \frac{H}{4} \times \frac{W}{4}}$ with suitable dimensions.

\textbf{Context Network.} Its architecture remains consistent with the feature network, and it adds a series of residual blocks and two additional downsampling layers, obtaining multi-level context features $\mathbf{f}^{c}_{i}$ ($i$ = 1, 2, 3) at 1/4, 1/8, 1/16 resolutions. Then we can get the initial hidden and the context information:
\begin{equation}
\begin{aligned}
    \mathbf{h}_{i} = & \tanh(\mathbf{f}_{i}^{c}) \\
    \mathbf{c}_{i} = & \text{ReLU}(\mathbf{f}_{i}^{c})
\end{aligned}
\end{equation}

\begin{figure*}[t]
    \centering
    \includegraphics[width=1.0\textwidth]{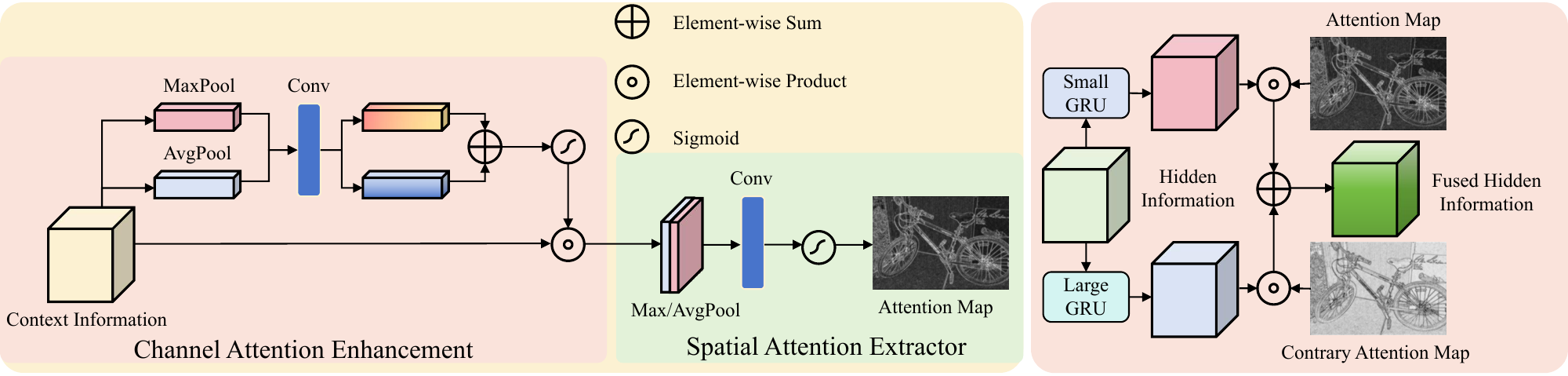}
    \caption{The architecture of proposed modules. Left: Contextual Spatial Attention (CSA) module. Right: Selective Recurrent Unit (SRU).}
    \label{fig:modules}
    \vspace{-10pt}
\end{figure*}

\subsection{Cost Volume Construction}

Given the left and the right features $\mathbf{f}, \mathbf{g}$, we first construct an all-pairs correlate cost volume:
\begin{equation}
    \mathbf{C}_{ijk} = \sum_{h} \mathbf{f}_{hij} \cdot \mathbf{g}_{hik}, \mathbf{C} \in \mathbb{R}^{\frac{H}{4} \times \frac{W}{4} \times \frac{W}{4}}
\end{equation}
Then we construct a 4-level correlation pyramid $\{\mathbf{C}_{i}\}$ ($i$ = 1, 2, 3, 4) by using 1D average pooling with a kernel size of 2 and a stride of 2 at the last dimension.

\subsection{Contextual Spatial Attention Module}
\label{sec:csa}

To help information from different receptive fields and frequencies fuse, the Contextual Spatial Attention (CSA) module extracts multi-level attention maps from context information as guidence. As illustrated in Fig. \ref{fig:modules}, CSA can be divided into two submodules: Channel Attention Enhancement (CAE) and Spatial Attention Extractor (SAE). These submodules are derived from CBAM~\cite{woo2018cbam} and we simplify them to better adapt to stereo matching.

\textbf{Channel Attention Enhancement.} Given a context information map $\mathbf{c} \in \mathbb{R}^{C \times H \times W}$, we first use an average-pooling and a max-pooling operation on the spatial dimension to get two maps $\mathbf{f}_{avg}, \mathbf{f}_{max} \in \mathbb{R}^{C \times 1 \times 1}$. Then we use two convolutional layers to perform feature transformation on these maps separably. After that, we add these two maps together and use the sigmoid function to convert them into weights $\mathbf{M}_{c} \in \mathbb{R}^{C \times 1 \times 1}$ between 0 and 1. Lastly, using an element-wise product, the initial map can capture which channel map has high feature values to be enhanced, and which channel map has low feature values to be suppressed.

\textbf{Spatial Attention Extractor.} After the CAE module, we continue to use the same pooling operations, but now we pool on the channel dimension. Then we concatenate these pooling maps to form a map in $\mathbb{R}^{2 \times H \times W}$ and use one convolutional layer with a sigmoid function to generate the final attention map. Reviewing previous operations, this attention map has high weights in regions needing high-frequency information because this information possesses high feature values in the context information. Similarly, it has low weights in regions needing low-frequency information. In general, the attention map can explicitly distinguish regions that need information at different frequencies.

\subsection{Selective Recurrent Unit}

To capture information at different frequencies, Selective Recurrent Unit (SRU) uses attention maps extracted by CSA to fuse hidden information derived from GRUs with different kernel sizes.

\begin{figure}
    \centering
    \includegraphics[width=0.4\textwidth]{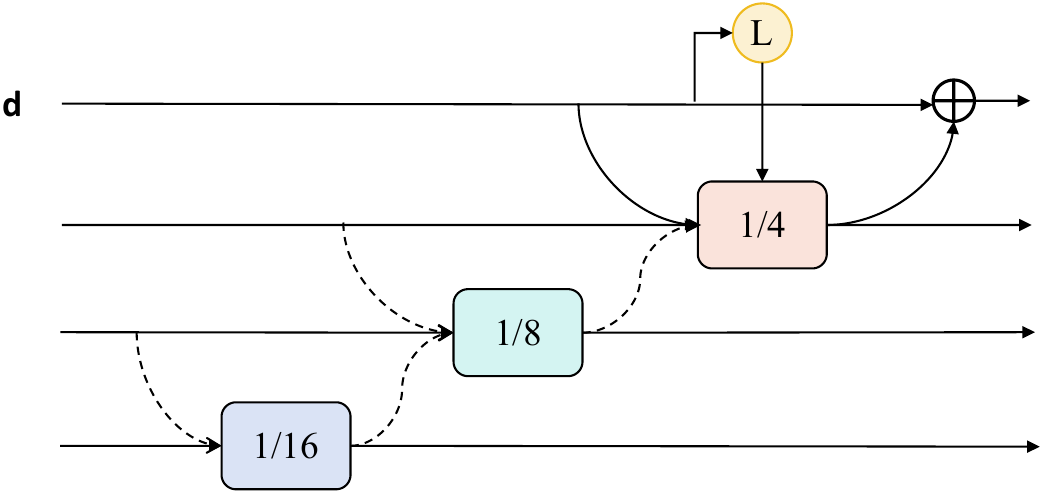}
    \caption{Multi-level SRU. Information is passed between SRUs at adjacent resolutions. Dashed arrows represent upsampling and downsampling operations. At $1/4$ resolution, disparity and local cost volume will be additional information put into SRUs.}
    \label{fig:multi-level}
    \vspace{-10pt}
\end{figure}

\textbf{Multi-level update structure.} As illustrated in Fig. \ref{fig:multi-level}, SRUs at $1/8$, $1/16$ resolutions take the attention map, context information, hidden information at the same resolution, and the hidden information at adjacent resolutions as inputs. At $1/4$ resolution, SRUs take disparity, and local cost volume as additional inputs, and then their outputs will go through two convolutional layers to generate disparity residuals. The local cost volume is derived from the all-pairs correlation pyramid in the same way as RAFT-Stereo~\cite{lipson2021raft}. At last, disparities at $1/4$ resolution will be upsampled into full resolution using the convex combination.

\textbf{SRU's architecture}. A single GRU can be defined as follows:
\begin{equation}
\begin{aligned}
    z_{k} = & \sigma(\text{Conv}([h_{k-1}, x_{k}], W_z)), \\
    r_{k} = & \sigma(\text{Conv}([h_{k-1}, x_{k}], W_r)), \\
    \Tilde{h}_{k} = & \tanh(\text{Conv}([r_{k} \odot h_{k-1}, x_{k}], W_h)), \\
    h_{k} = & (1 - z_{k}) \odot h_{k-1} + z_{k} \odot \Tilde{h}_{k}
\end{aligned}
\end{equation}
where $x_k$ is the concatenation of disparity, correlation, hidden information, and context information previously defined. Unlike RAFT-Stereo~\cite{lipson2021raft} that divide the context information into $c_z$, $c_r$, $c_h$, we add it into $x_k$ because using convolutions with different kernel sizes can fully utilize context information.

As illustrated in Fig. \ref{fig:modules}, a single SRU can be defined as follows:
\begin{equation}
    h_{k} = \mathbf{A} \odot h_{k}^{s} + (1 - \mathbf{A}) \odot h_{k}^{l}
\end{equation}
where $\mathbf{A}$ denotes the attention map derived from CSA at the same resolution, $h_{k}^{s}$ denotes the GRU with smaller kernel sizes and $h_{k}^{l}$ denotes the larger one. 

As Sec. \ref{sec:csa} mentioned, the attention map has high weights in regions needing high-frequency information. Therefore, the GRU with smaller kernel sizes that can capture high-frequency information like edge, and thin objects should do element-wise products with the attention map directly, and the GRU with larger kernel sizes should do element-wise products with the contrary attention map.

\textbf{Receptive fields analysis.}  The repective fields computing formula~\cite{araujo2019computing} can be defined as follows:
\begin{equation}
    r_{0} = \sum_{l=1}^{L}((k_{l} - 1)\prod_{i=1}^{l-1}s_{i}) + 1
\end{equation}
where $k_l$ denotes the kernel size, $s_i$ denotes the stride size, and $r_0$ denotes the whole network.

Given a multi-level structure like Fig. \ref{fig:multi-level}, if we take the $1/4$ resolution as the basis, and the downsampling operations can be regarded as a
convolution with kernel size 3, stride size 2, this structure's receptive fields are $k$, $2k + 3$, $3k + 6$. That means it only has $3$ fixed receptive fields in total.

If we replace GRUs with our SRUs with a small kernel size $s$ and a large kernel size $l$, the multi-level structure will have $6$ receptive fields initially. Besides, pixels in hidden information are affected by different receptive fields during fusion, and the fusion is influenced by attention maps adaptively. In general, the multi-level SRU holds dynamic receptive fields, and it enables itself to capture information at different frequencies.

\subsection{Loss Function}

We supervise our network on the L1 distance between all predicted disparities $\{\mathbf{d}_i\}^{N}_{i=1}$ and the ground truth disparity $\mathbf{d}_{gt}$ with increasing weights. The total loss is defined as:
\begin{equation}
    \mathcal{L} = \sum_{i=1}^N \gamma ^{N - i} ||\mathbf{d}_i - \mathbf{d}_{gt}||_1
\end{equation}
where $\gamma=0.9$, and $N$ is the number of iterations.
\section{Experiments}
\label{sec:exp}

\textbf{Scene Flow}~\cite{mayer2016large} is a synthetic dataset including 35,454 training pairs and 4,370 testing pairs with dense disparity maps. For training and testing, we use the finalpass version, because it contains more realistic and difficult effects than the cleanpass version. \textbf{KITTI 2012}~\cite{geiger2012we} and \textbf{KITTI 2015}~\cite{menze2015object} are datasets for real-world driving scenes. KITTI 2012 contains 194 training pairs and 195 testing pairs, and KITTI 2015 contains 200 training pairs and 200 testing pairs. \textbf{ETH3D}~\cite{schops2017multi} is a collection of gray-scale stereo pairs containing 27 training pairs and 20 testing pairs for indoor and outdoor scenes. \textbf{Middlebury}~\cite{scharstein2014high} is a high-resolution dataset containing 15 training pairs and 15 testing pairs for indoor scenes.

\subsection{Implementation Details}

We implement our Selective-Stereo with PyTorch and the model is trained on NVIDIA RTX 3090 GPUs. For all experiments, we use the AdamW~\cite{loshchilov2017decoupled} optimizer and clip gradients to the range [-1, 1]. We use the one-cycle learning rate schedule with a learning rate of 2e-4. We first train our model on Scene Flow with a batch size of 8 for 200k steps as the pretrained model. The crop size is $320 \times 720$, and we use 22 update iterations during training.

\subsection{Ablation Study}

In this section, we evaluate our model in different settings to verify our proposed modules in several aspects. All results use 32 update iterations.

\textbf{Effectiveness of proposed modules}. To verify the effectiveness of our proposed modules, we take RAFT-Stereo~\cite{lipson2021raft} as the baseline and replace its GRUs with our SRUs. As shown in Tab. \ref{tab:effectiveness}, the proposed SRU can improve the accuracy even without CSA. It means that if we just sum up the information from different branches, the growth of receptive fields can be beneficial for inference. If we add our CSA but invert the weights of the attention maps, the effect even decreases. That validates that our CSA's attention maps do indeed reflect the weights of information at different frequencies in regions. Therefore, if we add CSA normally, the full model (Selective-RAFT) can achieve the best performance with only a 4\% increase in parameters.

\begin{table*}[t]
\centering
\begin{tabular}{l|cccc|cc|c}
\hline
Model                      & GRU & SRU & \begin{tabular}[c]{@{}c@{}}CSA\\ (Contrary)\end{tabular} & CSA & \begin{tabular}[c]{@{}c@{}}EPE\\ (px)\end{tabular} & \begin{tabular}[c]{@{}c@{}}\textgreater{}1px\\ (\%)\end{tabular} & \begin{tabular}[c]{@{}c@{}}Param\\ (M)\end{tabular} \\ \hline
Baseline (RAFT-Stereo)               & \checkmark   &     &                                                          &     & 0.53                                               & 6.08                                                             & 11.12                                               \\ \hline
SRU                          &     & \checkmark   &                                                          &     & 0.50                                               & 5.38                                                             & 11.65                                               \\
SRU+CSA (Contrary)              &     & \checkmark   & \checkmark                                                        &     & 0.50                                               & 5.58                                                             & 11.65                                               \\
Full model (Selective-RAFT) &     & \checkmark   &                                                          & \checkmark   & \textbf{0.47}                                               & \textbf{5.32}                                                             & 11.65                                               \\ \hline
\end{tabular}
\caption{Ablation study of the effectiveness of proposed modules on the Scene Flow test set. SRU denotes Selective Recurrent Unit, and CSA denotes Contextual Spatial Attention. Contrary means we invert the weights of the attention maps. The baseline is RAFT-Stereo.}
\label{tab:effectiveness}
\end{table*}

\begin{figure*}[t]
    \centering
    \includegraphics[width=0.9\textwidth]{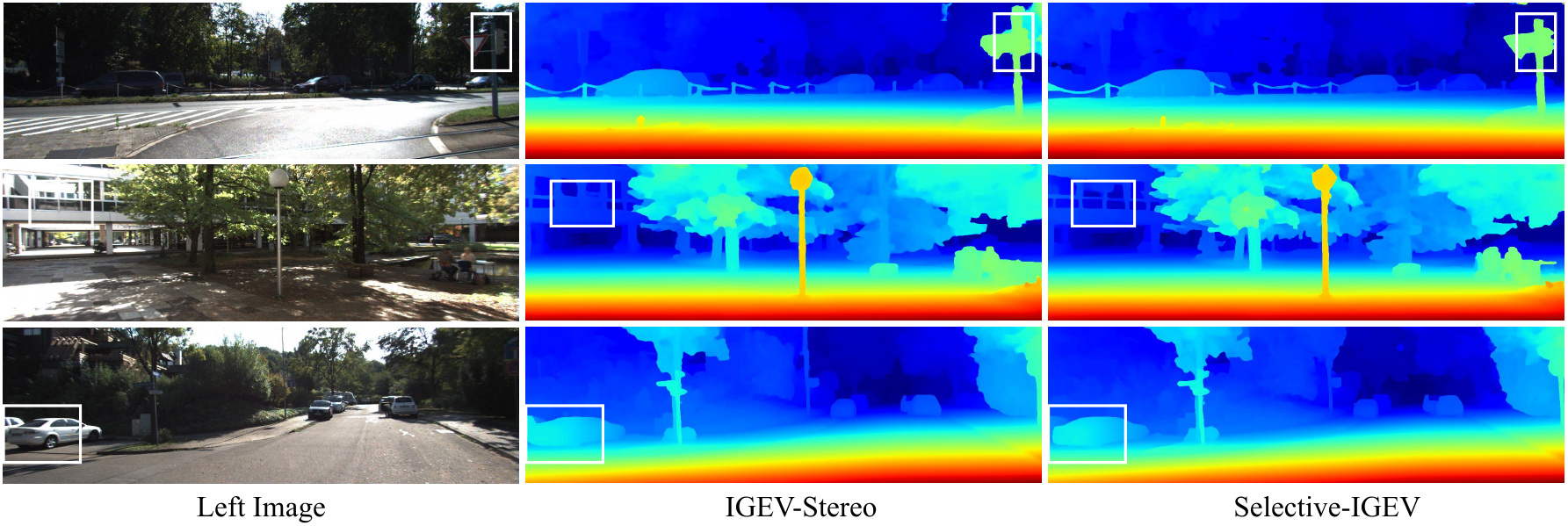}
    \caption{Qualitative results on the test set of KITTI. Our Selective-IGEV outperforms IGEV in detailed and weak texture regions.}
    \label{fig:kitti}
    \vspace{-10pt}
\end{figure*}

\begin{table} \footnotesize
\centering
\begin{tabular}{lccc}
\hline
Model          & EPE (px)      & \textgreater{}1px (\%) & Param (M) \\ \hline
RAFT-Stereo~\cite{lipson2021raft}    & 0.53          & 6.08                   & 11.12     \\
Selective-RAFT & \textbf{0.47} & \textbf{5.32}          & 11.65     \\ \hline
IGEV-Stereo~\cite{xu2023iterative}    & 0.47          & 5.21                   & 12.60     \\
Selective-IGEV & \textbf{0.44} & \textbf{4.98}          & 13.14     \\ \hline
DLNR~\cite{zhao2023high}           & 0.49          & 5.06                   & 57.37     \\
Selective-DLNR & \textbf{0.46} & \textbf{4.73}          & 58.09     \\ \hline
\end{tabular}
\vspace{-5pt}
\caption{Ablation study of the universality of proposed modules.}
\label{tab:universality}
\vspace{-5pt}
\end{table}

\begin{table} \footnotesize
\centering
\begin{tabular}{lcccccc}
\hline
\multirow{2}{*}{Model} & \multicolumn{6}{c}{Number of Iterations} \\ \cline{2-7} 
                       & 1     & 2    & 3    & 4    & 8    & 32   \\ \hline
RAFT-Stereo~\cite{lipson2021raft}            & 2.08  & 1.13 & 0.87 & 0.75 & 0.58 & 0.53 \\
Selective-RAFT         & \textbf{1.95}  & \textbf{1.06} & \textbf{0.81} & \textbf{0.69} & \textbf{0.53} & \textbf{0.47} \\ \hline
IGEV-Stereo~\cite{xu2023iterative}            & 0.66  & 0.62 & 0.58 & 0.55 & 0.50 & 0.47 \\
Selective-IGEV         & \textbf{0.65}  & \textbf{0.60} & \textbf{0.56} & \textbf{0.53} & \textbf{0.48} & \textbf{0.44} \\ \hline
\end{tabular}
\vspace{-5pt}
\caption{Ablation study of the number of iterations.}
\label{tab:number}
\vspace{-5pt}
\end{table}

\begin{table}
\centering
\begin{tabular}{lcc}
\hline
Kernel Sizes & EPE (px) & \textgreater{}1px (\%) \\ \hline
$1 \times 1 + 1 \times 5$      & 0.48     & 5.41                   \\
$3 \times 3 + 1 \times 5$      & 0.48     & \textbf{5.30}                   \\
$1 \times 1 + 3 \times 3$      & \textbf{0.47}     & 5.32                   \\ \hline
\end{tabular}
\caption{Ablation study of the size of convolutional kernels.}
\label{tab:size}
\vspace{-10pt}
\end{table}

\begin{figure}
    \centering
    \includegraphics[width=1.0\linewidth]{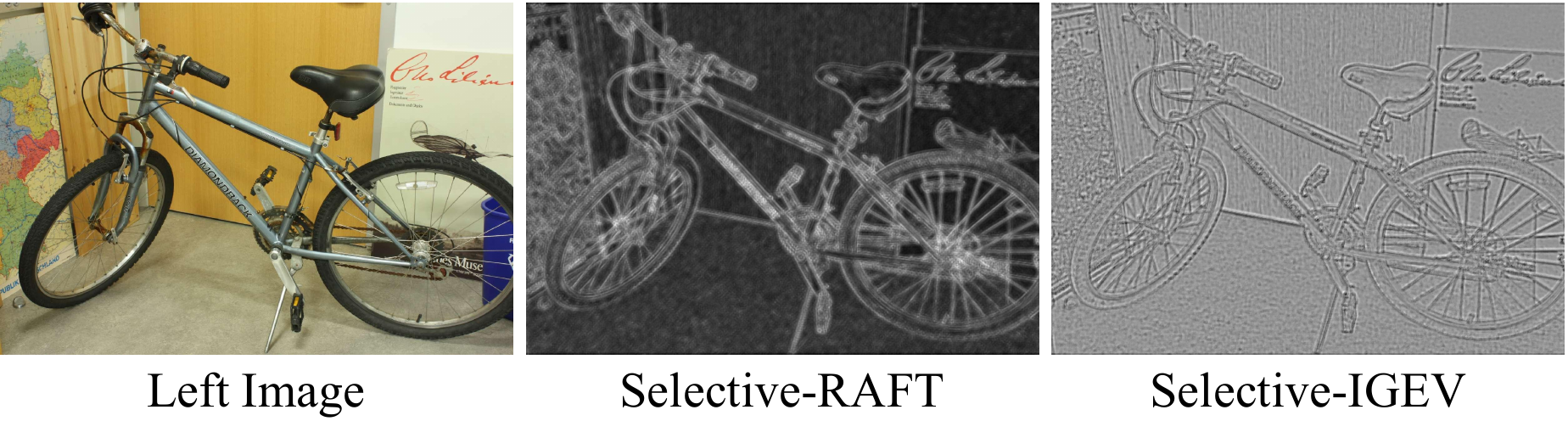}
    \caption{Visualization of the attention map of different networks.}
    \label{fig:attention}
    \vspace{-10pt}
\end{figure}

\begin{table*}[t] \footnotesize
\centering
\begin{tabular}{lccccccc}
\hline
Method   & CSPN & LEAStereo & LaC + GANet & ACVNet & IGEV-Stereo & Selective-RAFT (Ours) & Selective-IGEV (Ours) \\ \hline
EPE (px) & 0.78 & 0.78      & 0.72        & 0.48   & 0.47        & 0.47                  & \textbf{0.44}         \\ \hline
\end{tabular}
\vspace{-5pt}
\caption{Quantitative evaluation on Scene Flow test set.}
\label{tab:sceneflow}
\vspace{-5pt}
\end{table*}

\begin{table*}[t]
\centering
\begin{tabular}{l|cccccc|ccc|c}
\hline
\multirow{2}{*}{Method} & \multicolumn{6}{c|}{KITTI 2012}                                                             & \multicolumn{3}{c|}{KITTI 2015}               & \multirow{2}{*}{\begin{tabular}[c]{@{}c@{}}Run-time\\ (s)\end{tabular}} \\
                        & 2-noc         & 2-all         & 3-noc         & 3-all         & EPE-noc      & EPE-all      & D1-bg         & D1-fg         & D1-all        &                                                                         \\ \hline
AcfNet~\cite{zhang2020adaptive}                  & 1.83          & 2.35          & 1.17          & 1.54          & 0.5          & 0.5          & 1.51          & 3.80          & 1.89          & 0.48                                                                    \\
LEAStereo~\cite{cheng2020hierarchical}               & 1.90          & 2.39          & 1.13          & 1.45          & 0.5          & 0.5          & 1.40          & 2.91          & 1.65          & 0.30                                                                    \\
ACVNet~\cite{xu2022attention}                  & 1.83          & 2.35          & 1.13          & 1.47          & 0.4 & 0.5          & 1.37          & 3.07          & 1.65          & 0.20                                                                    \\
RAFT-Stereo~\cite{lipson2021raft}             & 1.92          & 2.42          & 1.30          & 1.66          & 0.4 & 0.5          & 1.58          & 3.05          & 1.82          & 0.38                                                                    \\
PCWNet~\cite{shen2022pcw}                  & 1.69          & 2.18          & \textbf{1.04} & \textbf{1.37} & 0.4 & 0.5          & 1.58          & 3.05          & 1.82          & 0.38                                                                    \\
LaC + GANet~\cite{liu2022local}             & 1.72          & 2.26          & 1.05          & 1.42          & 0.4 & 0.5          & 1.44          & 2.83          & 1.67          & 1.80                                                                    \\
CREStereo~\cite{li2022practical}               & 1.72          & 2.18          & 1.14          & 1.46          & 0.4 & 0.5          & 1.45          & 2.86          & 1.69          & 0.41                                                                    \\
IGEV-Stereo~\cite{xu2023iterative}             & 1.71          & 2.17          & 1.12          & 1.44          & 0.4 & 0.4 & 1.38          & 2.67          & 1.59          & 0.18                                                                    \\
Selective-RAFT (Ours)          & 1.64          & 2.09          & 1.10          & 1.43          & 0.4 & 0.5          & 1.41          & 2.71          & 1.63          & 0.45                                                                    \\
Selective-IGEV (Ours)          & \textbf{1.59} & \textbf{2.05} & 1.07          & 1.38          & 0.4 & 0.4 & \textbf{1.33} & \textbf{2.61} & \textbf{1.55} & 0.24                                                                    \\ \hline
\end{tabular}
\caption{Quantitative evaluation on KITTI 2012 and KITTI 2015.}
\label{tab:kitti}
\vspace{-10pt}
\end{table*}

\begin{table}
\centering
\begin{tabular}{lcccc}
\hline
\multirow{2}{*}{Method} & \multicolumn{2}{c}{Edges}              & \multicolumn{2}{c}{Non-Edges}          \\ \cline{2-5} 
                        & EPE      & \textgreater{}1px & EPE      & \textgreater{}1px \\ \hline
RAFT-Stereo~\cite{lipson2021raft}             & 3.21          & 29.16                  & 0.53          & 6.53                   \\
Selective-RAFT          & \textbf{2.40} & \textbf{21.63}         & \textbf{0.40} & \textbf{4.65}          \\ \hline
IGEV-Stereo~\cite{xu2023iterative}             & 2.23          & 20.42                  & 0.41          & 4.58                   \\
Selective-IGEV          & \textbf{2.18} & \textbf{20.01}         & \textbf{0.38} & \textbf{4.35}          \\ \hline
\end{tabular}
\caption{Quantitative evaluation on Scene Flow test set in different regions.}
\label{tab:edge}
\vspace{-10pt}
\end{table}

\textbf{Universality of proposed modules.} To verify the universality of our proposed modules, we take three typical iterative stereo matching methods as the baseline and replace their GRUs with SRUs. Especially, in DLNR~\cite{zhao2023high}, the recurrent units are LSTMs but not GRUs, so we just replace GRUs inside SRUs with LSTMs to make a fair comparison. As shown in Tab. \ref{tab:universality}, all methods have a significant improvement in the EPE metrics on Scene Flow, and the insertion of modules only results in a slight increase in parameters. Besides, as shown in Fig. \ref{fig:attention}, the CSA module generates different attention maps in different networks. In Selective-RAFT, because the cost volume contains a large amount of noisy information, the network needs more large kernels to filter the local cost volume. On the contrary, the cost volume has already been aggregated in Selective-IGEV, so the network tends to maintain high-frequency information using small kernels. Moreover, the cost volume in Selective-IGEV faces an over-smooth problem~\cite{xu2023iterative}, and that's why the attention map tends to increase the weights of large kernels to recover edge regions. In general, the CSA module shows different tendencies in different networks, which is a reflection of its adaptive ability. 

\textbf{Number of iterations.} Our Selective-Stereo can achieve better performance with a smaller number of iterations. As shown in Tab. \ref{tab:number}, our Selective-RAFT get the same performance with only 8 iterations compared to RAFT-Stereo~\cite{lipson2021raft}, and for IGEV-Stereo~\cite{xu2023iterative}, our Selective-IGEV also get a slight improvement with a few iterations. It shows that our modules can make secondary filtering to reduce noisy information from the initial cost volume.

\textbf{Size of convolution kernels.} We verify different kernel sizes on Scene Flow as shown in Tab. \ref{tab:size}. At last, we choose the combination of $1 \times 1$ and $3 \times 3$ as our default configuration, because it achieves a competitive performance and reduces computational costs.

\subsection{Comparisons with State-of-the-art}

All fine-tuned models use the model pretrained on Scene Flow. Different target datasets use different finetune strategies. We validate two models called Selective-RAFT and Selective-IGEV using RAFT-Stereo~\cite{lipson2021raft} and IGEV-Stereo~\cite{xu2023iterative} as the baseline respectively.

\textbf{Scene Flow.} As shown in Tab. \ref{tab:sceneflow}, we achieve a new state-of-the-art EPE of 0.44 on Scene Flow with Selective-IGEV, which surpasses LaC + GANet~\cite{liu2022local} by 38.89\%. Besides, our Selective-RAFT also achieves a competitive EPE of 0.47 compared to IGEV-Stereo~\cite{xu2023iterative} with smaller parameters. To validate the ability to fuse information by regions of our modules, we then split Scene Flow test set into two regions: edge regions and non-edge regions using the Canny operator. As shown in Tab. \ref{tab:edge}, our Selective-RAFT outperforms RAFT-Stereo~\cite{lipson2021raft} by 25.23\% and 24.53\% in edge regions and non-edge regions. Due to IGEV-Stereo's aggregated cost volume~\cite{xu2023iterative}, there's only a slight improvement in edge regions, but our Selective-IGEV still outperforms it by 7.32\% in non-edge regions.

\begin{table*}[t]
\centering
\begin{tabular}{l|cccc|cccc}
\hline
\multirow{2}{*}{Method} & \multicolumn{4}{c|}{ETH3D}                                    & \multicolumn{4}{c}{Middlebury}                                \\
                        & Bad 1.0       & Bad 0.5       & Bad 4.0       & AvgErr        & Bad 2.0       & Bad 1.0       & Bad 4.0       & AvgErr        \\ \hline
CroCo-Stereo~\cite{weinzaepfel2023croco}            & 0.99          & 3.27          & 0.13          & 0.14          & 7.29          & 16.9          & 4.18          & 1.76          \\
GMStereo~\cite{xu2023unifying}                & 1.83          & 5.94          & 0.08          & 0.19          & 7.14          & 23.6          & 2.96          & 1.31          \\
HITNet~\cite{tankovich2021hitnet}                  & 2.79          & 7.83          & 0.19          & 0.20          & 6.46          & 13.3          & 3.81          & 1.71          \\
IGEV-Stereo~\cite{xu2023iterative}             & 1.12          & 3.52          & 0.11          & 0.14          & 4.83          & 9.41          & 3.33          & 2.89          \\
RAFT-Stereo~\cite{lipson2021raft}             & 2.44          & 7.04          & 0.15          & 0.18          & 4.74          & 9.37          & 2.75          & 1.27          \\
CREStereo~\cite{li2022practical}               & \textbf{0.98} & 3.58          & 0.10          & 0.13          & 3.71          & 8.25          & 2.04          & 1.15          \\
EAI-Stereo~\cite{zhao2022eai}              & 2.31          & 5.21          & 0.70          & 0.21          & 3.68          & 7.81          & 2.14          & 1.09          \\
DLNR~\cite{zhao2023high}                    & -             & -             & -             & -             & 3.20          & 6.82          & 1.89          & 1.06          \\
Selective-RAFT (Ours)   & 1.69          & 5.78          & 0.13          & 0.17          & -             & -             & -             & -             \\
Selective-IGEV (Ours)   & 1.23          & \textbf{3.06} & \textbf{0.05} & \textbf{0.12} & \textbf{2.51} & \textbf{6.53} & \textbf{1.36} & \textbf{0.91} \\ \hline
\end{tabular}
\caption{Quantitative evaluation on ETH3D and Middlebury benchmarks. Note: Middlebury only allows one publish per paper, so we only publish our Selective-IGEV.}
\label{tab:eth3d-middlebury}
\vspace{-10pt}
\end{table*}

\begin{figure*}[t]
    \centering
    \includegraphics[width=0.85\textwidth]{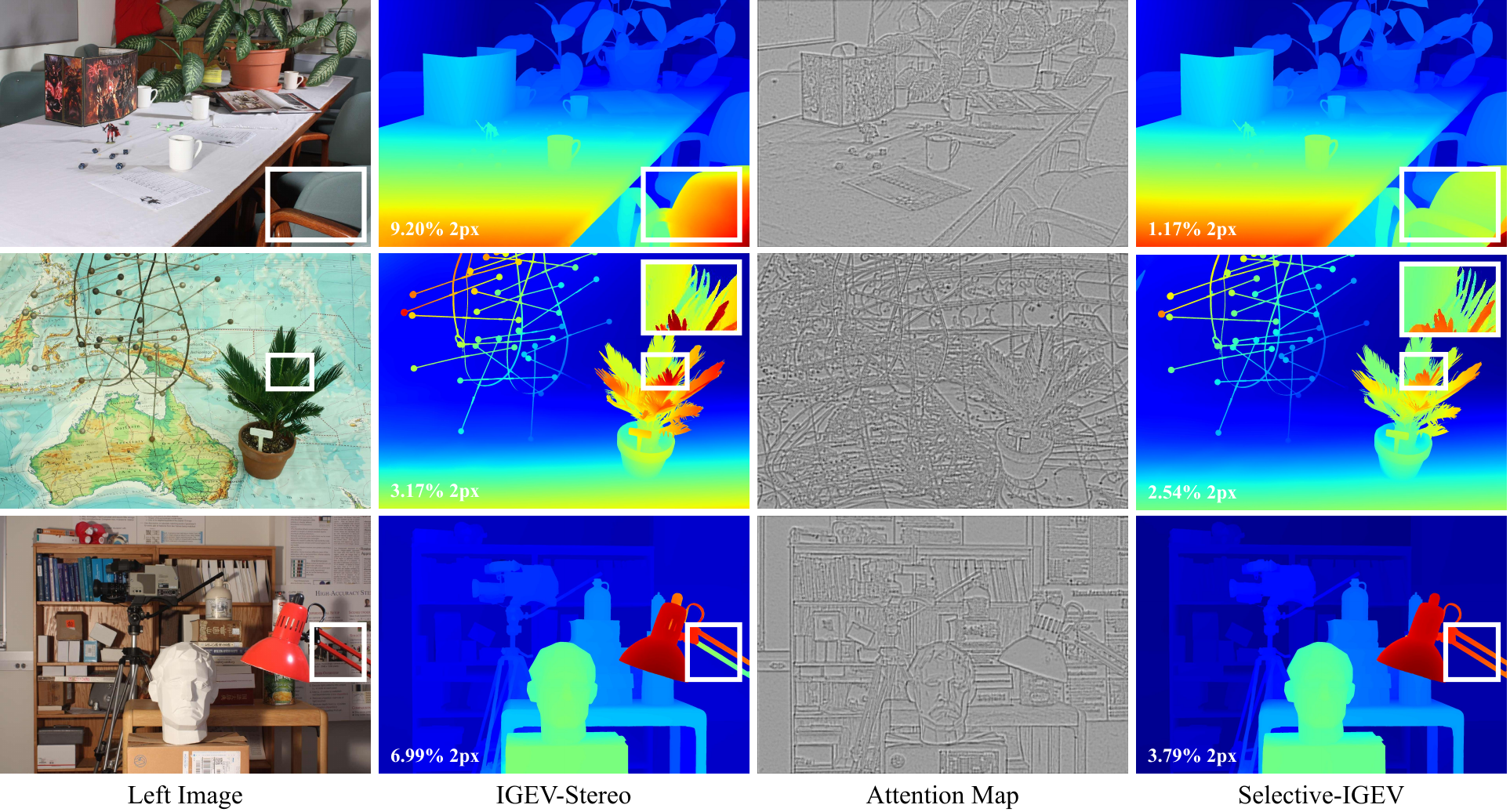}
    \caption{Qualitative results on the test set of Middlebury. The third column is the visualization of attention maps generated by the CSA module. Our Selective-IGEV outperforms IGEV in large textureless and thin object regions.}
    \label{fig:middlebury}
    \vspace{-10pt}
\end{figure*}

\textbf{KITTI.} We finetune our model on the mixed dataset of KITTI 2012 and KITTI 2015 with a batch size of 8 for 50k steps. Then we evaluate our Selective-Stereo on the test set of KITTI 2012 and KITTI 2015. As shown in Tab. \ref{tab:kitti}, we achieve the best performance among all published methods for almost all metrics. On KITTI 2012, our Selective-RAFT outperforms RAFT-Stereo~\cite{lipson2021raft} by 14.58\% and 13.64\% on 2-noc and 2-all metrics, and our Selective-IGEV ranks $1^{st}$ on these metrics. On KITTI 2015, our Selective-RAFT outperforms RAFT-Stereo~\cite{lipson2021raft} by 10.44\% on the D1-all metric, and our Selective-IGEV ranks $1^{st}$ on all metrics with only 16 iterations same as IGEV-Stereo~\cite{xu2023iterative}. As shown in Fig. \ref{fig:kitti}, our Selective-IGEV outperforms IGEV-Stereo~\cite{xu2023iterative} in detailed and textureless regions.

\textbf{ETH3D.} Following CREStereo~\cite{li2022practical} and GMStereo~\cite{xu2023unifying}, we use a collection of several public stereo datasets for training. The crop size is $384 \times 512$ and we first finetune the Scene Flow pretrained model on the mixed Tartan Air~\cite{wang2020tartanair}, CREStereo Dataset~\cite{li2022practical}, Scene Flow~\cite{mayer2016large}, Sintel Stereo~\cite{butler2012naturalistic}, InStereo2k~\cite{bao2020instereo2k} and ETH3D~\cite{schops2017multi} datasets for 300k steps. Then we finetune it on the mixed CREStereo Dataset~\cite{li2022practical}, InStereo2k~\cite{bao2020instereo2k} and ETH3D~\cite{schops2017multi} datasets with for another 90k steps. As shown in Tab. \ref{tab:eth3d-middlebury}, our Selective-RAFT outperforms RAFT-Stereo~\cite{lipson2021raft} by 17.90\% on Bad 0.5 metric, and our Selective-IGEV achieves the best performance among all published methods for almost all metrics. 

\textbf{Middlebury.} Also following CREStereo~\cite{li2022practical} and GMStereo~\cite{xu2023unifying}, we first finetune the Scene Flow pretrained model on the mixed Tartan Air~\cite{wang2020tartanair}, CREStereo Dataset~\cite{li2022practical}, Scene Flow~\cite{mayer2016large}, Falling Things~\cite{tremblay2018falling}, InStereo2k~\cite{bao2020instereo2k}, CARLA HR-VS~\cite{yang2019hierarchical} and Middlebury~\cite{scharstein2014high} datasets using a crop size of $384 \times 512$ for 200k steps. Then we finetune it on the mixed CREStereo Dataset~\cite{li2022practical}, Falling Things~\cite{tremblay2018falling}, InStereo2k~\cite{bao2020instereo2k}, CARLA HR-VS~\cite{yang2019hierarchical} and Middlebury~\cite{scharstein2014high} datasets using a crop size of $384 \times 768$ with a batch size of 8 for another 100k steps. As shown in Tab. \ref{tab:eth3d-middlebury}, our Selective-IGEV achieves the best performance among all published methods. As shown in Fig \ref{fig:middlebury}, compared to IGEV-Stereo~\cite{xu2023iterative}, our Selective-IGEV performs better in textureless, and detailed regions. The third column in Fig \ref{fig:middlebury} is the visualization of attention maps generated by the CSA module. It shows that attention maps can surely split regions that require information at different frequencies.

\section{Conclusion}
\label{sec:con}

We propose Selective-Stereo, a novel iterative stereo matching method. The proposed Contextual Spatial Attention module and Selective Recurrent Unit help the network capture information at different frequencies for edge and smooth regions. Our Selective-Stereo ranks $1^{st}$ on KITTI, ETH3D, and Middlebury in almost all metrics among all published methods. It shows an ability to fuse information at different frequencies adaptively for edge and smooth regions with the help of attention maps extracted by CSA.

However, our method still faces some challenges. Firstly, although our method can fuse information adaptively using attention maps, the SRU's receptive field is still limited by predefined values. Secondly, adding branches or increasing the sizes of convolutional kernels leads to high memory and time costs, so we will explore the combination of lightweight convolutions and our method to reduce memory costs. Lastly, it's also a good direction to do research on the combination of convolutions and self-attention due to their different advantages and receptive fields.

\hspace*{\fill}

\noindent\textbf{Acknowledgement.} This research is supported by National Natural Science Foundation of China (62122029, 62061160490, U20B200007).

\newpage
{
    \small
    \bibliographystyle{ieeenat_fullname}
    \bibliography{main}
}

\clearpage
\setcounter{page}{1}
\maketitlesupplementary

\section{Results on Ill-posed Regions}

To demonstrate the ability of our method to handle ill-posed regions, we evaluate it on KITTI 2012 reflective regions.
As shown in Tab. \ref{tab:reflective}, our Selective-RAFT outperforms RAFT-Stereo~\cite{lipson2021raft} by almost 20\% and our Selective-IGEV ranks $1^{st}$ among all publish methods. It shows that our method can fuse information at different frequencies to overcome ill-posed regions. Specifically, large kernels capture global low-frequency information to tackle local ambiguities, while small kernels capture local high-frequency information to maintain detailed structures. When encountering reflective regions, weights of large kernels increase to fuse more global information, while the network still fuses local information captured by small kernels to maintain details. 

\begin{table} \footnotesize
\centering
\begin{tabular}{l|cccc}
\hline
\multirow{2}{*}{Method} & \multicolumn{4}{c}{KITTI 2012 (Reflective Regions)}            \\
                        & 2-noc         & 2-all         & 3-noc         & 3-all         \\ \hline
AcfNet~\cite{zhang2020adaptive}                  & 11.17         & 13.13         & 6.93          & 8.52          \\
LEAStereo~\cite{cheng2020hierarchical}               & 9.66          & 11.40         & 5.35          & 6.50          \\
ACVNet~\cite{xu2022accurate}                  & 11.42         & 13.53         & 7.03          & 8.67          \\
RAFT-Stereo~\cite{lipson2021raft}             & 8.41          & 9.87          & 5.40          & 6.48          \\
PCWNet~\cite{shen2022pcw}                  & 8.94          & 10.71         & 4.99          & 6.20          \\
LaC + GANet~\cite{liu2022local}             & 10.40         & 12.21         & 6.02          & 7.34          \\
CREStereo~\cite{li2022practical}               & 9.71          & 11.26         & 6.27          & 7.27          \\
IGEV-Stereo~\cite{xu2023iterative}             & 7.29          & 8.48          & \underline{4.11}    & 4.76          \\
Selective-RAFT (Ours)         & \underline{7.19}    & \underline{7.96}    & 4.35          & \underline{4.68}    \\
Selective-IGEV (Ours)          & \textbf{6.73} & \textbf{7.84} & \textbf{3.79} & \textbf{4.38} \\ \hline
\end{tabular}
\caption{Quantitative evaluation on KITTI 2012 reflective regions (ill-posed regions).}
\label{tab:reflective}
\vspace{-10px}
\end{table}

\end{document}